# Accuracy and Fidelity Comparison of Luna and DALL-E 2 Diffusion-Based Image Generation Systems

Michael Cahyadi
Computer Science Department
School of Computer Science
Bina Nusantara University
Jakarta, Indonesia
michael.cahyadi001@binus.ac.id

Muhammad Rafi
Computer Science Department
School of Computer Science
Bina Nusantara University
Jakarta, Indonesia
muhammad.rafi007@binus.ac.id

William Shan
Computer Science Department
School of Computer Science
Bina Nusantara University
Jakarta, Indonesia
william.sitanggang@binus.ac.id

Henry Lucky
Computer Science Department
School of Computer Science
Bina Nusantara University
Jakarta, Indonesia
henry.lucky@binus.ac.id

Jurike V. Moniaga
Computer Science Department
School of Computer Science
Bina Nusantara University
Jakarta, Indonesia
jurike@binus.edu

*Abstract* — We qualitatively examine the accuracy and fidelity between two diffusion-based image generation systems, namely DALL-E 2 and Luna, which have massive differences in training datasets, algorithmic approaches, prompt resolvement, and output upscaling. The methodology used is a qualitative benchmark created by Saharia et al. and in our research we conclude that DALL-E 2 significantly edges Luna in both alignment and fidelity comparisons.

## I. Introduction

Image generation systems is one of the many avenues artificial intelligence research projects have been pursuing ways to improve generative methods. Image generation systems have immense potential to compliment and extend human creativity [1], but on the other hand there are issues with the field such as potential abuse to spread misinformation and harrassment [2], bias against certain cultural groups [3], and harmful associations against marginalized societies [4].

The field of image generation using artificial intelligence technologies has made great advancements in the last few years, with recent models capable of generating images with near human-like characteristics. Variations of Generative Adversarial Networks (GAN) [5] were among the first models to generate high-quality images, but recently there has been more focus by researchers and the public [6] on diffusion-based models trained using massive datasets.

The open nature of research information regarding diffusion-based image generation models have also led to an increase of image generation systems made by individuals, which may not have sufficient guardrails against abuse [7].

Image generation systems also use CLIP latents to associate certain human concepts and understand them in creative contexts [8]. With artificial intelligence systems continuing to get better in non-analytical areas such as artistic creativity that mimic closely human cognitive architectures, researchers might soon get closer into the realm of General Artificial Intelligence (GAI) [9].

As artificial intelligence becomes a more pervaise tool in day-to-day workflows, there needs to be an evaluation regarding the quality of outputs generated by image generation systems. Accurately judging the alignment and perceived fidelity of generated outputs from these image generation systems can help researchers and developers build better systems that are aware of the pitfalls of current systems in the market.

While there exists many diffusion-based image generation systems, both open and closed sourced, we decided to search for two systems who adopt resolvement approaches that lead in their industry in terms of accuracy and widescale implementation in the image generation technology community. Latent diffusion models and CLIP-guided [10] diffusion models represent the forefront methodologies for image-generation technologies with alignment and fidelity results surpassing previous GAN-based systems.

This paper also ultimately aims to examine the difference in accuracy between images generated by diffusion-based systems that are made by a large company using a large training data set and a system created by an individual with more limited training resources and less guardrails towards abuse. To that effect, we consider the following two image generation models for comparing our results:

1. **DALL-E 2** [8] is an image generation system created by OpenAI which can generate high-resolution images that combine various concepts and art styles. The project was built in PyTorch using ViT-H/16 text encounters with the training data of 650M images scraped from the internet and aligned by CLIP [10].

2. **Luna** is an image generation system built by Arfy Slowly, a Senior Software Engineer at Google Research. The project was built in Tensorflow and published on GitHub as an open-source project. The system uses a latent diffusion model [11] to condition the model on text prompts, however the training dataset used is unknown.

## II. Related Works

There have been many papers that try to compare the performance of two image generation systems, whether quantitatively or qualitatively. The metrics that are used to verify the accuracy of image generation systems mostly rely on image fidelity and its benchmark against real-world equivalents.

While quantitative methods have been laid out to gauge the accuracy of image generation systems such as the Fréchet Inception Distance by Heusel et al. [12], the metric is used to compare GAN performance at image generation using real-life samples, which differs from our attempts to gauge the accuracy of image generation systems at prompt resolving novel concepts that have little to no real-life examples.

Existing evaluations of diffusion-based image generation models are mostly based on the accuracy of the image generated in specific fields such as in artificially generated faces [13]. But no paper has qualitatively evaluated the inherent accuracy between the prompt given to the model to the generated image.

Qualitative methods of performance analysis are usually done by human surveyors such as in research by Saharia et al. [14]. This is due to the subjective nature of art [15], unlike measurable things such as image fidelity, which cannot be measured with common metric calculations.

In [14], the method used in evaluating image accuracy consists of 2 questions given to human raters inquiring about the fidelity and alignment (the accuracy between the human interpretation of a given concept and the output given by an artificial intelligence [16]) of the system's output.

III. METHODOLOGY

A. Prompt Creation

The prompts listed below are modified prompts from Google's Drawbench Benchmark [14], which covers a variety of concepts, art styles, and common pitfalls of image generation systems to generate images that can be a point of evaluation for the alignment and fidelity of the image generation systems. We have also listed the reasoning behind why we choose each prompt.

The prompts detailed above have been screened by running them through a Google image search and seeing how easily images for these concepts could be retrieved; from this process, we eliminated two prompts and modified another.

TABLE I. VARIOUS PROMPTS USED

| Number | Contents | |
|---|---|---|
| | Prompt | Explanation |
| 1 | A photorealistic image of a machine resembling a human being and able to replicate certain human movements and functions automatically. | Prompt is used to evaluate the ability of the image generation system to build photorealistic images that don't cross the uncanny valley [17]. |
| 2 | A half-robot and man entity with chainsaws for their head and hands is the style of Japanese anime. | Prompt is used to evaluate the bias in machine learning algorithms that are trained with data from westernized-culture [18]. |
| 3 | Rbefraigerator. | The prompt is used as a way to determine the image generation system's ability to navigate word misspellings [19]. |
| 4 | A car on top of a spoon. | The prompt is used to examine the ability of image generation systems to generate images novel in concept [19]. |
| 5 | Two bicycles and one car on an empty grass field. | The prompt is used to examine the ability of image generation systems to generate images with accurate positional information [19]. |
| 6 | In late afternoon in January in Jakarta, a man stands in the shadow of a tree. | The prompt is used to examine the ability of image generation systems to accurately create shadows that correspond with differing lighting conditions [19] |
| 7 | A Sumatran tiger under the sea. | The prompt is used to examine the ability of the image generation system to generate images that have conflicting concepts [19]. |
| 8 | Art nouveau stained glass window art depicting Woody from Toy Story. | The prompt is used to examine the ability of the image generation system to generate images with pop culture products in a medium not usually associated with the product [20]. |

B. Image Generation

Below each set of 4 images per-prompt on each model, we will outline general observations from the researchers as with the cited reasons of why each model behave in such a way. The image results are compiled for further analysis in the paper in methodologies outlined in later subsections.

The researchers will generate every 32 images from each system, noting that DALL-E 2 generates four images per run. For DALL-E 2, access was provided to the system during September 2022 after a request for beta-testing research was approved by the company. DALL-E 2 was accessed from the OpenAI Beta website, with 8 credits dispensed every month for non-commercial research use only. DALL-E 2 outputs 4 photos in one-prompt execution. DALL-E 2 outputs 1024x1024 pixel images and Luna outputs 512x512 pixel images.

For Luna, we use the provided Colab notebook by Google to run the system. Considerations we're made to run Luna using on-premises hardware, however due to Tensorflow's requirement of NVIDIA CUDA cores we decided to opt for cloud solutions instead due to faster compute times and as a better benchmark against DALL-E 2 which is a cloud-based system hosted in Microsoft Azure. Luna outputs 4 photos in one-prompt execution.

## C. Analysis

As laid out in the related works section, due to art being inherently subjective in nature [15], normal metrics cannot be applied when analyzing the inherent accuracy of prompt creations from image generation systems.

The methodologies to evaluate these images are based on Drawbench by [14] who used human raters to judge prompt accuracy of Imagen, Google's in-house proprietary image generation system, against existing competitors such as OpenAI DALL-E 2 [8] and GLIDE [21].

For the benchmark analysis, we conduct an independent human evaluation run for each category. For each prompt, the rater is shown two sets of images with one from DALL-E 2, and second from Luna. Each set contains eight non-cherry-picked generations from the corresponding model. The human rater will be asked two questions.

1. **Which set of images better represents the text caption: [Text Caption]?** Question subjectively evaluates image-text alignment.

2. **Which set of images is of higher quality?** Question subjectively evaluates image fidelity.

For each question, the rater is asked to select from two choices:

1. **I prefer set A.**
2. **I prefer set B.**

The paper aggregates the scores from different raters and then score it in a percentage value which will be presented in the form of a candle graph. The authors did not perform any post filtering of the data to identify unreliable raters, both for expedience of the analysis process and because the task was straightforward to explain and execute.

## IV. RESULTS

After carefully compiling the results of the survey from human raters over a span of two weeks. We analyze the results according to generally acceptable benchmarks for alignment and fidelity scores.

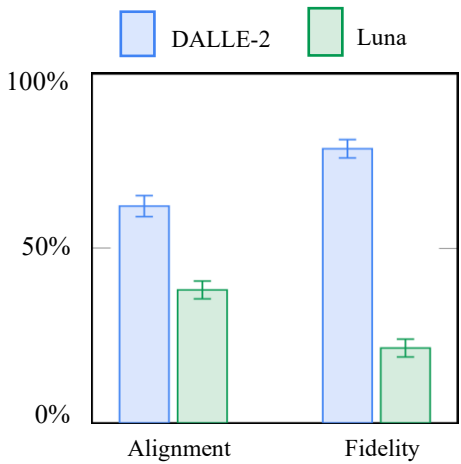

Fig. 1. Alignment and Fidelity comparison between DALL-E 2 and Luna using methodologies outlined in [14]. plotted into a candle graph: User preference rates for prompt alignment and image fidelity.

Results show that when the output images are given to human raters and evaluated using methods outlined in [14], the results show that DALL-E 2 on average received a higher image-text alignment (62.1%) and image fidelity (83.4%) rating than Luna.

FID scores can be a more objective measurement of fidelity of machine-generated images, but previous research has shown that FID scores are not reflective of perceptual quality[22].

TABLE II. DALL-E 2 VS. LDM-KL FID COMPARISON

| Model | FID-30K | $N_{Params}$ |
|---|---|---|
| DALL-E 2 | 10.39 | 650M |
| Luna (LDM-KL) | 12.63 | n/a |

While quantitative measurements are outside the scope of this paper, FID scores cited from research papers of the respective models show that DALL-E 2 outperforms other methods on MS-COCO 256 x 256 with zero-shot FID-30K with a score of 10.39, significantly outperforming systems based on Latent Diffusion Models (LDM-KL) such as Luna. The results line up with human raters' indication of individual samples fidelity ratings.

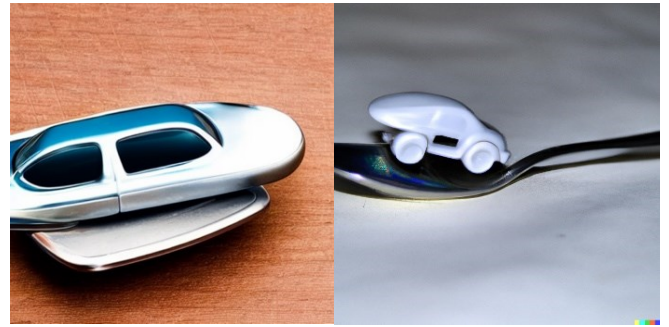

Fig. 2. Selected image samples from the resolvement process of Prompt 4 by Luna (left) and DALL-E 2 (right). Images we're picked from a set of 4 each generated per system.

It's observed that both prompt systems have difficulties in prompt resolvement of novel concepts, such as a car on a spoon. While Luna seems to struggle with the concept, DALL-E 2 interpret it as a request for a toy car, and not a real car.

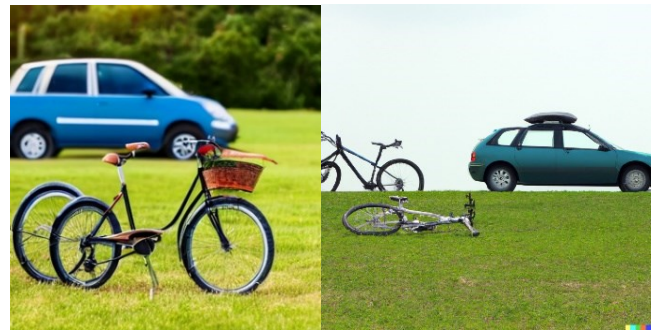

Fig. 3. Selected image samples from the resolvement process of Prompt 5 by Luna (left) and DALL-E 2 (right). Images we're picked from a set of 4 each generated per system.

It's also observed that Luna has difficulty assigning the correct number of items in an image given a prompt that contains numerical amounting values. While the issue is also present in DALL-E 2, prior research has proven that the system can atleast count to four objects [19].

Researchers behind DALL-E 2 has also disclosed issues regarding compositionality [8], which is the ability to comprehend the merging of multiple object properties such as shape and positioning within the image. Which is why the placement of the objects inside of the picture generated might look too symetrical.

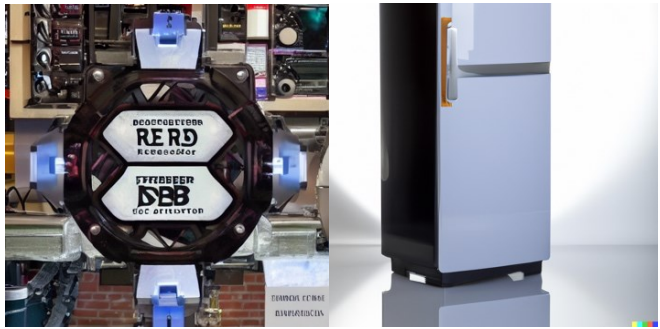

Fig. 4. Selected image samples from the resolvement process of Prompt 3 by Luna (left) and DALL-E 2 (right). Images we're picked from a set of 4 each generated per system.

Resolvement of misspelled prompts [19] has also proved a challenge for LDM-based systems such as Luna with DALL-E 2 accurately representing the misspelled prompt as a "refrigerator" and Luna failing to generate a comprehensible image. This is theorized to be the result of significantly better prompt alignment within DALL-E 2's generation system that enables it to edge out Luna in this prompt.

Resolvement of misspelled prompts [19] has also proved a challenge for LDM-based systems such as Luna with DALL-E 2 accurately representing the misspelled prompt as a "refrigerator" and Luna failing to generate a comprehensible image. This is theorized to be the result of significantly better prompt alignment within DALL-E 2's generation system that enables it to edge out Luna in this prompt.

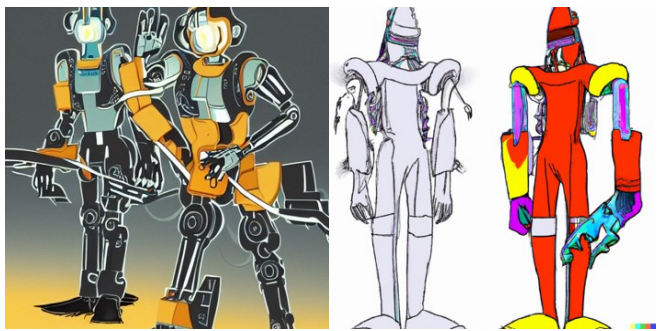

Fig. 5. Selected image samples from the resolvement process of Prompt 2 by Luna (left) and DALL-E 2 (right). Images we're picked from a set of 4 each generated per system.

Resolvement of prompts with non-westernized artstyles both failed to generate anything resembling the inputted prompt. Machine learning systems have consistently hit difficulties in detecting and generating styles that are uncommon outside of western culture such as the artstyle of Japanese anime [23] This is possibly the result of bias within large compiled datasets that's mainly trained on webscrapes of mostly western-aligned content [24]. This challenge will also present itself more in bigger datasets, which will complicate efforts to effectively scale computer vision and generative datasets without significant alignment.

The possibility of training differences affecting the performance of image generation systems are also observed to be correlated. Comparing FID-30K scores and observing the interception distance of FID-2K scores between the two systems have yielded interesting technical observations. The figures show that Luna experiences a distinct lowered amount of TPU (Tensor Processing Units) training days compared to DALL-E 2, which can negatively impact the alignment quality and perceived fidelity of the image as less itterations are performed within a specific timeframe. Luna as an individually built system also possibly suffered from time limitations during training.

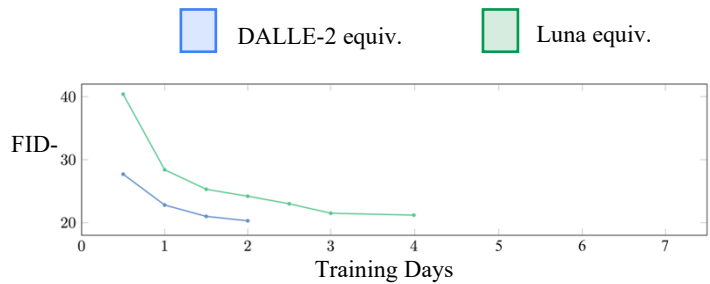

Fig. 6. Comparison of TPU training time needed to achieve a 20 basis point FID-2K rating between regular U-Nets (Luna or LDM-KL) vs efficient U-Nets (DALLE-2).

The differences mainly come down to complexity, which might have caused worse FID-2K scores due the amount of time used to train LDM-KL based systems compared to OpenAI's approach with DALL-E 2 [25]. Time differences may be attributed to the difference in libraries used, as Luna uses Tensorflow and DALL-E 2 uses PyTorch, the latter of which has been shown to be more performant than the former resulting in faster compute times [26].

## V. CONCLUSIONS

The round of experimentation showcases the effectiveness of frozen large pretrained language models as text encoders for the text-to-image generation, but differences exist between the capabilities of large models such as DALL-E 2 and smaller scale models such as Luna.

Dramatically increasing the size of these language models have significantly more impact than scaling the U-Net size on overall performance on alignment and fidelity. This encourages future research directions on exploring even bigger language models as text encoders, both by companies and individuals.

But increasing datasets has also several kinks other than purely technical complications as there are ethical challenges relating to large datasets used for the image generation systems, particularly regarding subject data awareness and consent [27], [28] and some datasents even reflect stereotypes, offensive viewpoints, and derogatory associations of various marginalized identity groups [24].

While Luna was edged out in both alignment and fidelity measurements both in qualitative benchmarks through human raters and quantitative benchmarks through zero-shot FID-2K and FID-30K scores, it has reached a remarkable level of accuracy for a system that is built by an individual and trained using a limited dataset.

We also find considerable performance penalties incurred by Luna's use of Tensorflow compared to DALL-E 2's use of PyTorch which resulted in a slower comparative TPU training days compared to the latter, which affects training accuracy.

We ultimately conclude that while differences exist between large systems made by corporations and smaller individual made systems, the advent of diffusion-based image generation systems have lowered the barrier to enter the image generation field significantly. The advancement in research of generative AI technologies need to be paired with safeguards and acknowledgement of ethical concerns, working towards a safer implementation of systems.

## Acknowledgment

We give thanks to Arfy Slowy from the Google Brain Research Team in Singapore and Imre Bard from the OpenAI Alignment Research team for helping early discussions, and providing many helpful comments and suggestions throughout the project. We thank you the team at Kaggle and OpenAI for the free tiers given for testing and exploratory research purposes. Special thanks to Agneta Viola for reviewing grammatical and linguistical errors. We thank Herendra Kurniawan for their consistent and critical help with TPU resource allocation and Kaggle notebook initialization.